\documentclass[letterpaper, 10 pt, journal, twoside]{IEEEtran}
\usepackage{cite}

\ifCLASSINFOpdf
  \usepackage[pdftex]{graphicx}
\else
\fi
\usepackage{amsmath}

\usepackage{multirow}

\usepackage{booktabs}

\begin{document}
%
\title{Attitude-Estimation-Free GNSS and IMU Integration}
%
%
%

\author{Taro Suzuki$^{1}$%
\thanks{Manuscript received: August 14, 2023; Revised: October 30, 2023; Accepted: December 3, 2023.}
\thanks{This paper was recommended for publication by Editor Sven Behnke upon evaluation of the Associate Editor and Reviewers' comments.}
\thanks{$^{1}$Taro Suzuki is with Future Robotics Technology Center, Chiba Institute of Technology, Japan
        {\tt\footnotesize taro@furo.org}}%
\thanks{Digital Object Identifier (DOI): see top of this page.}
}
%
%

\markboth{IEEE Robotics and Automation Letters. Preprint Version. Accepted December, 2023}
{Suzuki: Attitude-Estimation-Free GNSS and IMU Integration} 

%



\maketitle

\begin{abstract}
   A global navigation satellite system (GNSS) is a sensor that can acquire 3D position and velocity in an earth-fixed coordinate system and is widely used for outdoor position estimation of robots and vehicles. Various GNSS/inertial measurement unit (IMU) integration methods have been proposed to improve the accuracy and availability of GNSS positioning. However, all these methods require the addition of a 3D attitude to the estimated state to fuse the IMU data. In this study, we propose a new optimization-based positioning method for combining GNSS and IMU that does not require attitude estimation. The proposed method uses two types of constraints: one is a constraint between states using only the magnitude of the 3D acceleration observed by an accelerometer, and the other is a constraint on the angle between the velocity vectors using the angular change measured by a gyroscope. The evaluation results with the simulation data show that the proposed method maintains the position estimation accuracy even when the IMU mounting position error increases and improves the accuracy when the GNSS observations contain multipath errors or missing data. The proposed method could improve positioning accuracy in experiments using IMUs acquired in real environments.
\end{abstract}


%
\IEEEpeerreviewmaketitle

%
%
%
%

\section{Introduction}
\IEEEPARstart{G}{lobal} navigation satellite systems (GNSS), represented by global positioning systems, have become an indispensable infrastructure in today's society to estimate locations in outdoor environments. For example, it plays an important role in various applications, such as the navigation of people and vehicles in cities, unmanned delivery robots, autonomous flight of drones, and automated driving of vehicles. However, GNSS alone is difficult to use for positioning in environments such as tunnels, under trees, and elevated structures, where signals from GNSS satellites are blocked. In urban environments, GNSS signals enter the antenna after reflection or diffraction. This phenomenon, which is known as multipath, considerably reduces positioning accuracy \cite{GNSS_General,gnss_understanding}. Consequently, studies have been conducted to improve the availability and accuracy of the GNSS \cite{gnss_handbook}.

The availability and accuracy of GNSS positioning in urban environments can be improved by combining 3D velocity computed from GNSS Doppler shift measurements. The velocity from Doppler measurements can be estimated with centimeter-level accuracy in an ideal open-sky environment, which is considerably higher than that calculated from the GNSS pseudorange measurements \cite{vel1}. In addition, Doppler measurements are more robust than the GNSS pseudorange because of its resistance to multipath error and instantaneous signal blocking \cite{vel4}. However, when there is prolonged signal blocking, or when only reflected signals are incident on the antenna, the combined Doppler velocity is not expected to improve the accuracy of the estimated position.

To solve this problem, several studies have combined the GNSS and inertial measurement units (IMUs) \cite{gpsins}. Acceleration and angular velocity IMU measurements are independent of the environment for measurement accuracy. Therefore, GNSS and IMU are complementary, and their combination improves the accuracy and availability of position estimation. Filtering methods, such as complementary filters \cite{madgwick} and Kalman filters (KFs) \cite{gpsins}, and more recently, optimization-based methods, such as graph optimization \cite{go_suyrvey}, are often used to combine GNSS and IMU.

However, GNSS/IMU integration has the following limitations
\begin{itemize}
   \item IMU integration requires 3D attitude estimation, and even in applications where only the position is required, the 6-DOF pose estimation problem must be solved.
   \item The position and attitude of the IMU relative to the vehicle frame must be rigorously measured or calibrated in advance. Any error in the IMU mounting position will degrade the position estimation accuracy after GNSS integration.
\end{itemize}

This paper proposes a new GNSS/IMU integration method that does not require attitude estimation. We propose a state-to-state constraint using only the magnitude of the acceleration vector of the IMU output and a velocity vector direction constraint by integrating the angular velocities. Because the proposed method does not estimate the 3D attitude, the effect of the IMU mounting angle and position on the accuracy is small and calibration is not necessary. Therefore, the proposed method is effective for problems where sensors (e.g., smartphones) containing GNSS and IMU are mounted on a vehicle or robot in different positions and orientations each time and used for navigation. It can also simplify the system for navigation applications in which only the position is required. 

\subsection{Related Studies}
Combining GNSS with other sensors has been extensively studied. Although GNSS combined with a camera \cite{go_gv,go_gvi1} and lidar \cite{go_gil1,go_gil2} have been proposed, it is most commonly combined with an IMU \cite{gpsins,gnssins_survey}. This study envisions a low-cost MEMS IMU consisting of a 3-axis accelerometer and a 3-axis gyroscope, combined with GNSS. Compared with GNSS, IMU has a high sampling rate (typically 100–200 Hz) and provides highly continuous acceleration and angular velocity measurements. Although the IMU can measure the velocity and angular changes by integrating its measurements, integration errors accumulate. Therefore, by combining IMU with GNSS, which has a low sampling rate but provides absolute position and velocity, error accumulation can be eliminated.

Various methods have been proposed for the combination of GNSS and IMU, the most common being the use of KF \cite{gpsins_kf1,gpsins_kf2}. Complementary-filter-based algorithms are frequently used in environments with limited computational resources \cite{madgwick}. Recently, GNSS/IMU integration using optimization methods has also been studied \cite{go_gi1,go_gi2,TDCP2,go_gi4,go_gi5}. In \cite{go_gi2,go_gi4}, a method for integrating GNSS pseudorange observations and IMU with tight coupling was proposed using pose graph optimization. In \cite{go_gi2,go_gi4}, in addition to GNSS and IMU, time differences in the carrier phase were integrated as precise velocity constraints. Compared with the method using KFs, the method using optimization is advantageous in terms of accuracy \cite{go_gi1,go_gi_noatt1}.

However, all these methods add a 3D attitude to the state to be estimated. To decompose and apply the 3D acceleration measurements from the accelerometer to the navigation coordinate system, the 3D attitude must be estimated. In Ref. \cite{go_gi_noatt1}, the IMU was integrated without including the attitude as a state; however, a 3D attitude was obtained directly from the attitude heading reference system (AHRS). Therefore, errors in the AHRS affected the accuracy of the position estimation. Instead of estimating the attitude, Ref.\cite{go_gi_noatt2} improved the accuracy of the position estimate using the zero-velocity observation at the constraint obtained from the IMU. However, this constraint is only applicable to stationary vehicles.

In addition, the combination of GNSS and IMU requires a homogeneous transformation matrix from the IMU coordinate system to the vehicle coordinate system (center of rotation of the vehicle) \cite{ins_alignment1}. In many cases, the IMU mounting angle and position with respect to the vehicle frame are accurately measured in advance, or the IMU mounting error is added to the estimated state and simultaneously estimated \cite{ins_alignment2}. When the IMUs are frequently removed and replaced, prior measurements of the IMU mounting angles and positions are time consuming.

\subsection{Contributions}
The contributions of the proposed method are as follows:
\begin{itemize}
   \item A new IMU constraint that does not require attitude estimation is proposed, and a combined GNSS/IMU method that is independent of the IMU mounting angle and position is developed.
   \item The proposed method significantly improved position estimation accuracy in environments where GNSS observations were unavailable or contained errors owing to satellite shielding or multipath.
   \item We used simulations and real-world experiments to compare the accuracy of the proposed method with that of the conventional 6-DOF pose estimation method, which adds attitude to the estimated state. We clarified the applicability and effectiveness of the proposed method.
\end{itemize}

\begin{figure}[t]
   \centering
   \includegraphics[width=85mm]{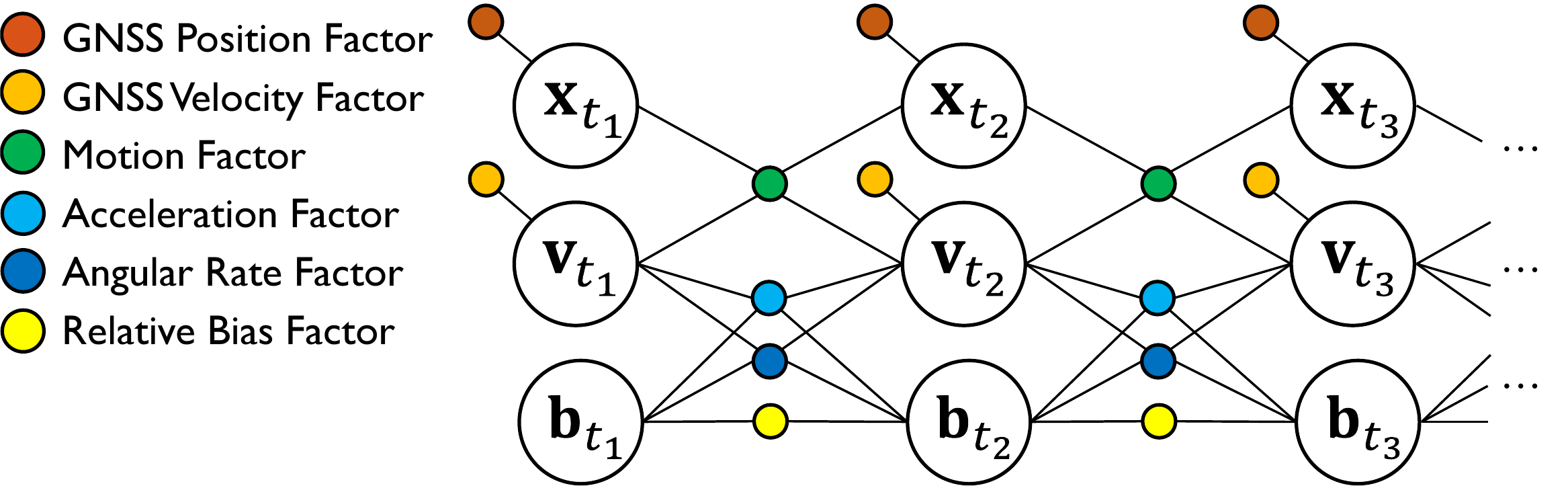} 
   \caption{Structure of the factor graph of the proposed method. The states are the 3D position $\mathbf{x}$, velocity $\mathbf{v}$, and IMU bias $\mathbf{b}$. The attitude is not included in the states. The IMU acceleration and angular velocity factors generate a constraint between successive velocities.}
   \label{fig1}
\end{figure}

\section{Proposed Method}
\subsection{Problem Setup}
In general, the 3D position $\mathbf{x}$ and 3D attitude ${\Phi}$ are used as the estimated states in a combined GNSS and IMU system. In addition, the 3D velocity $\mathbf{v}$ was estimated from the acceleration and angular velocity measured by the IMU. As low-cost IMUs possess large acceleration and angular velocity bias errors and drifts, the biases of each of the three axes, namely, $\mathbf{b}_{\mathrm{acc}}$ and $\mathbf{b}_{\mathrm{gyro}}$, are often added to the state. In a general GNSS/IMU integration, the estimated state at the $i$th epoch is as follows:

\begin{equation}
   \mathbf{X}_i=\left[\begin{array}{lllll}
      \mathbf{x}_i & {\Phi}_i & \mathbf{v}_i & \mathbf{b}_{\mathrm{acc},i} & \mathbf{b}_{\mathrm{gyro},i}
   \end{array}\right]^T
\end{equation}

The local east-north-up (ENU) coordinate system was adopted as the world frame. This is a 15-dimensional state-estimation problem. The measured values of the acceleration $\tilde{\mathbf{a}}$ and angular velocity $\tilde{\mathbf{\omega}}$ of the IMU in the IMU frame are as follows:

\begin{equation}
   \tilde{\mathbf{a}_i} = \mathbf{a}_i - \mathbf{C}_{W,i}^I \mathbf{g} +\mathbf{\omega}_i \times \left( \mathbf{\omega}_i \times \mathbf{r} \right) + \mathbf{b}_{\mathrm{acc},i}+\mathbf{\eta}_{\mathrm{acc}}
\end{equation}

\begin{equation}
   \tilde{\mathbf{\omega}_i} = \mathbf{\omega}_i+\mathbf{b}_{\mathrm{gyro},i}+\mathbf{\eta}_{\mathrm{gyro},i}
\end{equation}

\noindent where $\mathbf{a}$ is the translational acceleration measured in the IMU frame; $\mathbf{\omega}$ are the theoretical values of the angular velocity in IMU frame; $\mathbf{g}$ is the gravitational acceleration; $\mathbf{r}$ is the lever arm of the IMU from the center of rotation; $\eta$ represents Gaussian white noise; and $\mathbf{C}_W^I$ denotes the rotation matrix from the world frame to the IMU frame computed from the 3D attitude $\Phi$ and IMU mounting angle in the vehicle frame. 

The third term in (2) is the centripetal force, which is considerably large when the IMU is mounted on a vehicle or robot away from the center of rotation and generates a large centripetal force when turning. Accelerations other than gravitational acceleration act as disturbances when accelerometers are used to estimate the attitude. If a lever arm is present at the mounting position of the GNSS and IMU, the acceleration generated by the centripetal force significantly affects attitude estimation without prior information on the mounting position \cite{ins_alignment1}. To combine IMU observations with GNSS observations in the world frame, the 3D attitude $\Phi$ and IMU mounting angle, which represent the transformation from the IMU frame to the world frame, must be estimated. 

\subsection{Estimated State}
This study does not estimate the 3D attitude and IMU mounting angle and position; it combines the IMU and GNSS observations. The IMU and GNSS observations are exactly the same as those in the conventional method; however, the state has the following eight dimensions without attitude.

\begin{equation}
   \mathbf{X}_i=\left[\begin{array}{llll}
      \mathbf{x}_i & \mathbf{v}_i & \mathbf{b}_i
   \end{array}\right]^T
\end{equation}

\noindent where the 3D bias error of the accelerometer and gyroscope is estimated as the bias of the magnitude of the acceleration and the angular velocity in one dimension, respectively. 

\begin{equation}
   \mathbf{b}_i=\left[\begin{array}{ll}
      b_{\mathrm{acc},i} & b_{\mathrm{gyro},i}
   \end{array}\right]
\end{equation}


\subsection{Graph Structure}
In this study, factor graph optimization was used to estimate the position by combining the GNSS position and velocity, IMU acceleration, and angular velocity. A factor graph is a graphical representation of the constraints of a variable node, expressed at the edges between the variable and factor nodes \cite{gtsam}. 

The proposed method assumed that a GNSS and an IMU were attached to a vehicle; 3D position and velocity observations were obtained from the GNSS; and three-axis acceleration and angular velocity were obtained from the IMU. The 3D position and 3D velocity were calculated from the GNSS pseudorange and GNSS Doppler observation, respectively. Their accuracy was assumed to be a few meters for the position and 10 cm/s for the velocity as the accuracy of general GNSS. 

The graphical structure of the proposed method is shown in Fig. \ref{fig1}. The graph was constructed using the GNSS observation timing, which is a lower rate than IMU. The observation rate of GNSS was assumed to be 1–20 Hz, which is the output rate of general GNSS receivers, while that of the IMU was assumed to be 100–200 Hz. We proposed two new factors: (1) a constraint on the magnitude of the 3D acceleration vector, and (2) an angular velocity constraint between the velocity vectors based on the angular change. These IMU-based constraints were independent of the attitude and did not require the attitude to be added to the state. In addition, we used a motion factor that related the velocity to the position, GNSS position factor that was a 3D position constraint using GNSS pseudorange, and GNSS velocity factor that was a 3D velocity constraint using Doppler measurements.

\subsection{Acceleration Factor}
The acceleration factor constrained the change in successive 3D velocities using the magnitude of the 3D acceleration measurement from the accelerometer. Let $t_1$ and $t_2$ be the times of the $i$-th and $i+1$-th epochs, respectively. The error function of the acceleration factor is defined as

\begin{equation}
   e_{\mathrm{acc}, i}=\left\|\frac{\mathbf{v}_{i+1}-\mathbf{v}_{i}}{\Delta t_\mathrm{gnss}}-\mathbf{g}\right\|-\left\|\frac{\Delta t_{\mathrm{imu}}}{\Delta t_\mathrm{gnss}} \sum_{t=t_1}^{t_2} \mathbf{a}_t \right\|+b_{\mathrm{acc},i}
\end{equation}

\noindent where $\Delta t_\mathrm{gnss}$ and $\Delta t_\mathrm{imu}$ denote the observation time steps for GNSS and IMU, respectively. The first and second terms represent the magnitudes of the acceleration computed from the velocities and average acceleration between the states calculated from the accelerometers, respectively. This equation clearly indicates that the magnitude of the 3D acceleration vector can be used as a constraint to construct a constraint between states $\mathbf{v}_i$ and $\mathbf{v}_{i+1}$ that is independent of the attitude of the IMU. The total minimized error of the acceleration factor is calculated as follows:

\begin{equation}
   \left\|e_{\mathrm{acc}, i}\right\|_{\Omega_{\mathrm{acc}}}=e_{\mathrm{acc}, i} \, \Omega_{\mathrm{acc}} \, e_{\mathrm{acc}, i}
\end{equation}

\noindent where $\Omega_{\mathrm{acc}}$ is the information matrix, which is empirically determined from the catalog-specified noise model.

\subsection{Angular Velocity Factor}
The direction of the 3D velocity vector was constrained by the angular velocity observed by the gyroscope. This was expressed as a constraint that the angle formed by the velocity vectors between successive states coincided with the angular change resulting from the integration of the angular velocity measurements from the gyroscope. The error function of the angular velocity factor is defined as follows:

\begin{equation}
   e_{\mathrm{gyro}, i}=\arccos \left(\frac{\mathbf{v}_{i+1} \cdot \mathbf{v}_{i}}{\left\|\mathbf{v}_{i+1}\right\|\left\|\mathbf{v}_{i}\right\|}\right)-\left\|\sum_{t=t_1}^{t_2} \mathbf{\omega}_t \, \Delta t_{\mathrm{imu}}\right\|+b_{\mathrm{gyro},i}
\end{equation}

This equation defines the constraint that the angle between the velocity vectors is equal to the angular change owing to the integration of the angular velocity. The error function to be optimized is as follows:

\begin{equation}
   \left\|e_{\mathrm{gyro}, i}\right\|_{\Omega_{\mathrm{gyro}}}=e_{\mathrm{gyro}, i} \, \Omega_{\mathrm{gyro}} \, e_{\mathrm{gyro}, i}
\end{equation}

\noindent where $\Omega_{\mathrm{gyro}}$ is the information matrix, which is empirically determined from the specifications of the gyroscope and accelerometer. 

Unlike the acceleration factor, the angular velocity factor shown in Equation (8) requires that the orientation of the platform be aligned with the direction of the velocity vector. This is an important limitation that is applicable to general ground vehicles but not to differential-drive wheeled robots or quad-rotors. For these platforms, only the acceleration factor is used. In addition, this constraint is inapplicable when the velocity is zero or low. Therefore, this constraint is applied only when the magnitude of the velocity vector exceeds a certain threshold (1 m/s in this study).

\subsection{Other Factors}
\subsubsection{GNSS Position Factor}
The 3D positioning results from the least-squares method using GNSS pseudorange were combined for position estimation. The error function of the GNSS position factor is

\begin{equation}
   \mathbf{e}_{\mathrm{pos}, i}=\mathbf{x}_{i}-\mathbf{x}_{\mathrm{gnss},i}
\end{equation}

\noindent where $\mathbf{x}_{\mathrm{gnss},i}$ denotes the GNSS positioning solution in the ENU coordinate system. The information matrix of the GNSS position factor was computed from the covariance matrix estimated using the least-squares method. For the GNSS position error function, the M-estimator was applied as a robust optimization method owing to the existence of outliers caused by multipaths. The Huber function was used as the kernel \cite{HuberFGO}.

\subsubsection{GNSS Velocity Factor}
The 3D velocity was estimated using the least-squares method with GNSS Doppler shift measurements. As with the position factor, the 3D velocity $\mathbf{v}_{\mathrm{gnss},i}$ was converted to the ENU coordinate system and added to the graph as follows:

\begin{equation}
   \mathbf{e}_{\mathrm{vel}, i}=\mathbf{v}_{i}-\mathbf{v}_{\mathrm{gnss},i}
\end{equation}

Similar to the position factor, the covariance matrix estimated using the least-squares method was used to compute the information matrix. In general, the velocity calculated from the Doppler velocity is considerably more accurate than that calculated from the difference in pseudoranges. The M-estimator is used for the GNSS velocity error function as well as the GNSS position error function.

\subsubsection{Motion Factor}
The motion factor was used to relate the 3D velocity and position. Based on the average velocity and 3D position between successive states, the motion factor is defined as follows:

\begin{equation}
   \mathbf{e}_{\mathrm{m}, i}=\frac{\mathbf{x}_{i+1}-\mathbf{x}_i}{\Delta t_{\mathrm{gnss}}}-\frac{\mathbf{v}_{i+1}+\mathbf{v}_i}{2}
\end{equation}

The motion factor is a deterministic factor that relates the velocity and position. For the variance of its error, it uses a small fixed value that is adjusted heuristically. The acceleration and angular velocity constraints between the velocities are also propagated to the position using the motion factor, thereby improving the position estimation accuracy.

\subsubsection{Relative Bias Factor}
The relative bias factor between IMU bias states was used to control the variation in the acceleration and angular velocity biases. The following equations define the error functions for the relative bias factor:

\begin{equation}
   e_{\mathrm{bias,a}, i}=b_{\mathrm{acc},i+1}-b_{\mathrm{acc},i}
\end{equation}
\begin{equation}
   e_{\mathrm{bias,g}, i}=b_{\mathrm{gyro},i+1}-b_{\mathrm{gyro},i}
\end{equation}

\noindent where the information matrix of the relative bias factor is determined from the random-walk specifications of the accelerometer and gyroscope.

\subsection{Optimization}
The graph is constructed according to the factor described in the previous section using all GNSS and IMU observations in the dataset. In this study, we treated the optimization as a full-graph optimization problem using the Levenberg-Marquardt optimizer as a post-processing application. The final objective function to be optimized is expressed as follows:

\begin{IEEEeqnarray}{lCr}
   \widehat{\mathbf{X}}=\underset{\mathbf{X}}{\operatorname{argmin}} \left( \sum_{i} \left\|e_{\mathrm{acc}, i}\right\|_{\Omega_{\mathrm{acc}}}^{2}\right. +\sum_{i} \left\|e_{\mathrm{gyro}, i}\right\|_{\Omega_{\mathrm{gyro}}}^{2} \nonumber\\ 
   +\sum_{i}\left\|\mathbf{e}_{\mathrm{pos},i}\right\|_{\Omega_{\mathrm{pos},i}}^{2}
   +\sum_{i}\left\|\mathbf{e}_{\mathrm{vel},i}\right\|_{\Omega_{\mathrm{vel},i}}^{2}+\sum_{i}\left\|\mathbf{e}_{\mathrm{m},i}\right\|_{\Omega_{\mathrm{m}}}^{2} \nonumber\\
   +\sum_{i}\left\|e_{\mathrm{bias,a},i}\right\|_{\Omega_{\mathrm{bias,a}}}^{2}+\left. \sum_{i}\left\|e_{\mathrm{bias,g},i}\right\|_{\Omega_{\mathrm{bias,g}}}^{2} \right)
\end{IEEEeqnarray}

The proposed method was implemented using GTSAM, a general-purpose graph optimization library \cite{gtsam_github}. 

\section{Experiments}
We compared the proposed method with a conventional general 6-DOF GNSS/IMU integration method. To implement the 6-DOF GNSS/IMU integration method, we used a GNSS/IMU integration algorithm based on the IMU preintegration factor using GTSAM \cite{imu_preintegration1,imu_preintegration2}. The estimated state is a 15-dimensional vector as shown in (1). The proposed and comparison methods differed only in the graph structure of the IMU part; the difference lay in the use of the proposed acceleration factor and angular velocity factor or the conventional IMU pre-integration factor. We evaluated the proposed method using two datasets: simulation data and real data obtained by placing a smartphone on a vehicle.

\begin{figure}[t]
   \centering
   \includegraphics[width=75mm]{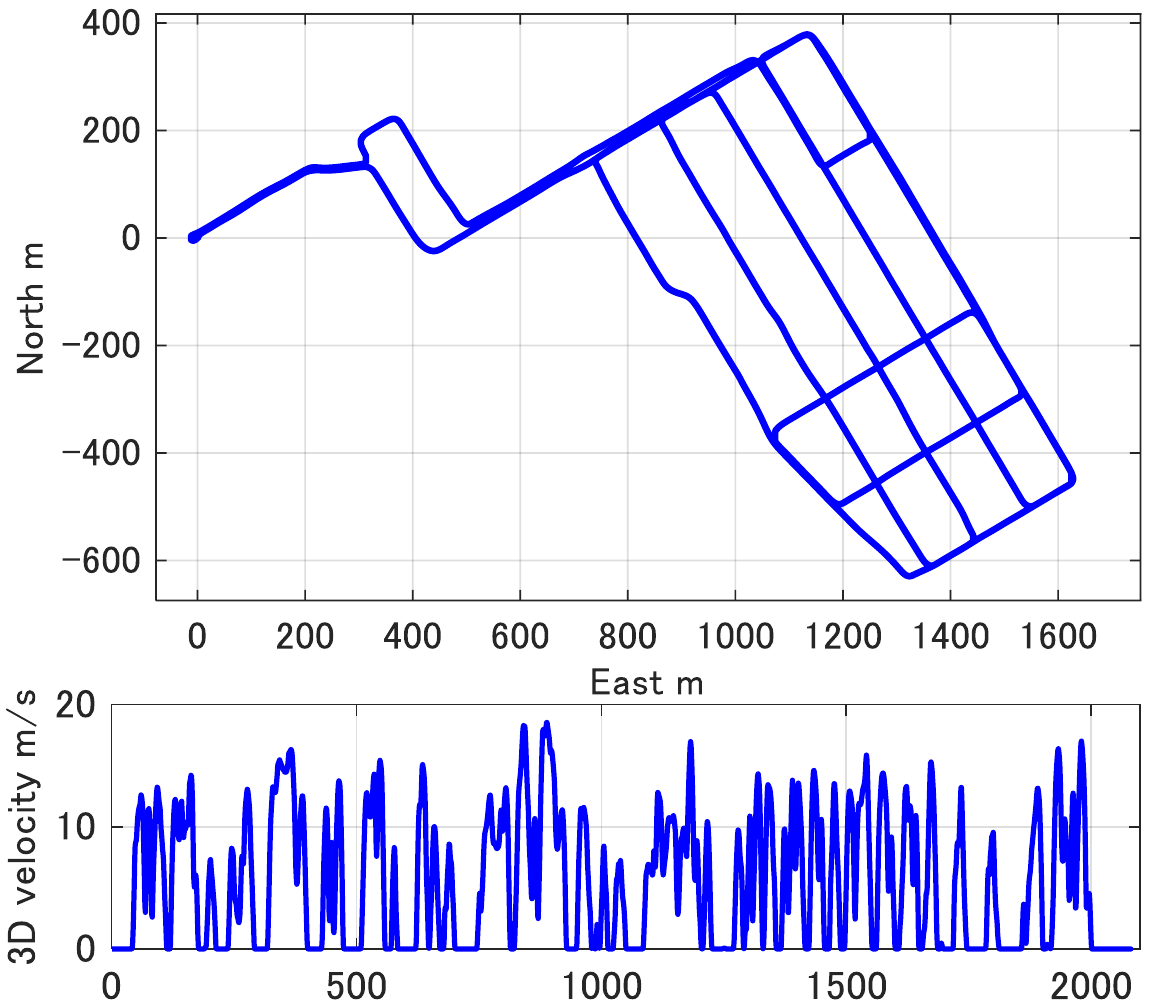} 
   \caption{Trajectory (top) and 3D velocity history (bottom) of the simulation data. The data was generated from the urban driving data of the vehicle.}
   \label{fig2}
\end{figure}

\begin{figure}[t]
   \centering
   \includegraphics[width=80mm]{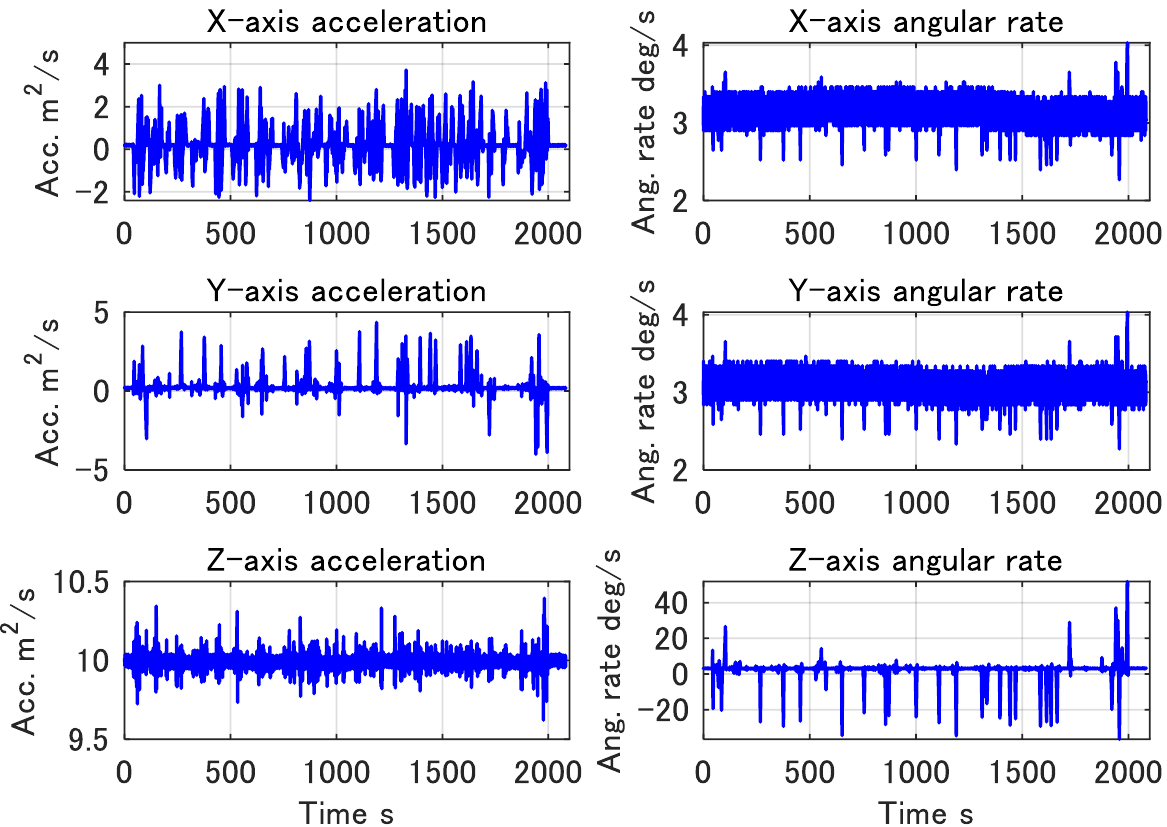} 
   \caption{IMU acceleration and angular velocity data generated from driving scenarios. The IMU mounting angles were matched to data from actual driving experiments.}
   \label{fig3}
\end{figure}

\begin{table}[]
   \centering
   \caption{Sensor parameters for IMU and GNSS data simulation.}
   \label{tab1}
   \begin{tabular}{@{}cccc@{}}
   \toprule
   Sensor & Parameter & Value & Unit \\ \midrule
   \multirow{4}{*}{Acceleration} & Noise density & $1.86\times 10^{-3}$ & $\mathrm{(m/s^2)/\sqrt Hz}$ \\
    & Random walk & $4.33 \times 10^{-4}$ & $\mathrm{(m/s^2)\sqrt Hz}$ \\
    & Constant bias & 0.19 & $\mathrm{m/s^2}$ \\
    & Sample rate & 100 & Hz \\ \midrule
   \multirow{4}{*}{Angular velocity} & Noise density & $1.87\times 10^{-4}$ & Unit \\
    & Random walk & $2.66\times 10^{-5}$ & $\mathrm{(rad/s)/\sqrt Hz}$ \\
    & Constant bias & 0.0545 & $\mathrm{(rad/s)\sqrt Hz}$ \\
    & Sample rate & 100 & rad/s \\ \midrule
   \multirow{2}{*}{GNSS position} & Noise density & 1.0 & m \\
    & Sample rate & 1 & Hz \\ \midrule
   \multirow{2}{*}{GNSS velocity} & Noise density & 0.2 & m/s \\
    & Sample rate & 1 & Hz \\ \bottomrule
   \end{tabular}%
 \end{table}

\subsection{Simulation Setup}
Simulation data of the IMU and GNSS observations were generated to evaluate the position-estimation performance of the proposed method. The MATLAB Navigation Toolbox was used to generate the simulation data for the IMU and GNSS. To generate sensor data, we used the vehicle trajectory from the Google Smartphone Decimeter Challenge \cite{gsd}, an evaluation using real data described below. 

Fig. \ref{fig2} shows the trajectory of the vehicle in the simulation data. The travel distance is approximately 12 km, including vehicle stops and starts in an urban environment, and takes approximately 35 min. 

Fig. \ref{fig3} shows the generated IMU measurements (acceleration and angular velocity). The parameters of the sensor models used to generate the simulation data are listed in Table I. Parameters such as the accelerometer, gyroscope noise density, and random walk were set using parameter estimates from low-cost IMUs based on \cite{imuparam}. GNSS observations were generated by adding Gaussian noise to the 3D position and 3D velocity. The noise of the GNSS velocity observations was set lower than that of the position observations. IMU measurements were generated at 100 Hz, and GNSS observations were generated at 1 Hz. 

Several simulation data points were generated with errors added to the IMU mounting position to compare the proposed and conventional methods. 
An AMD Ryzen 3950X @3.5GHz CPU was used for the evaluation experiments.

\begin{figure}[t]
   \centering
   \includegraphics[width=75mm]{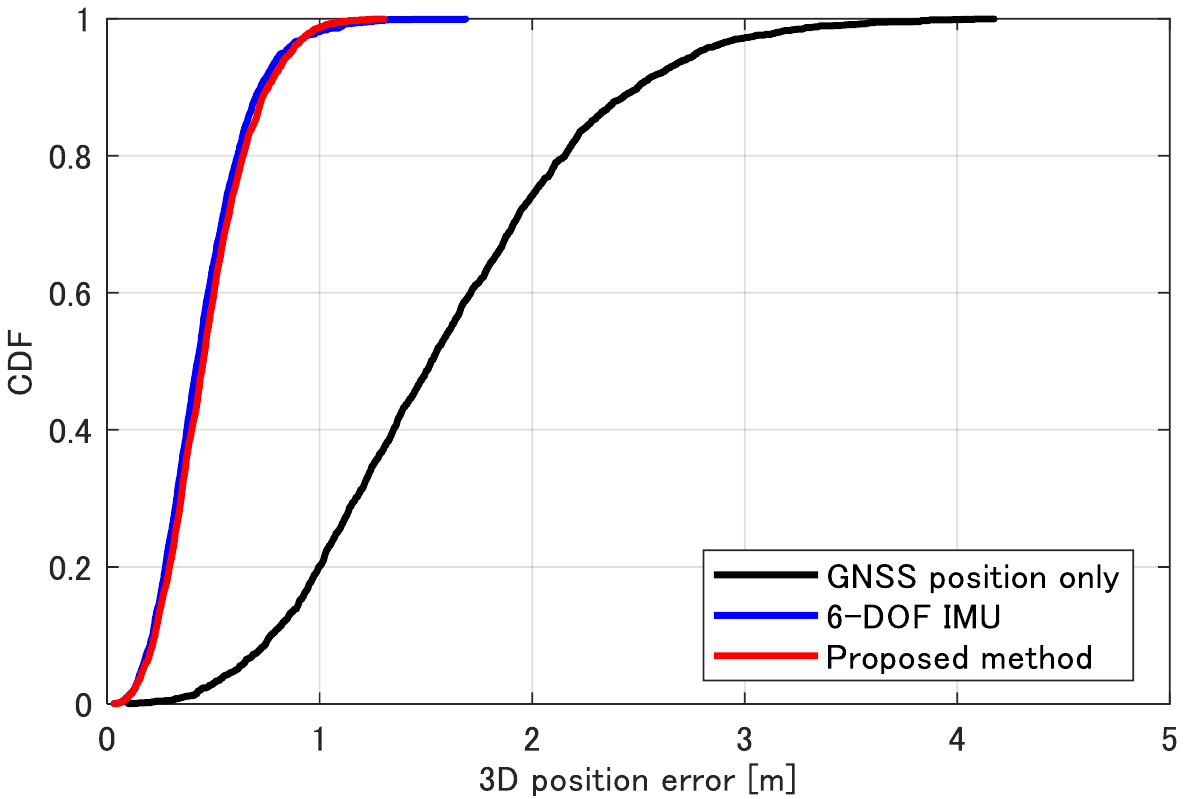} 
   \caption{Cumulative distribution function of the 3D position error for the proposed (red line) and conventional 6-DOF pose estimation (blue line). The combined IMU significantly improves the accuracy of position estimation in both cases.}
   \label{fig4}
\end{figure}

\begin{table}[]
   \centering
   \caption{RMS position error between the proposed method and conventional 6-DOF estimation using simulation data.}
   \label{tab2}
   \resizebox{\columnwidth}{!}{%
   \begin{tabular}{cccccccc}
   \hline
   \multicolumn{1}{l}{}                                       & \begin{tabular}[c]{@{}c@{}}East error\\  m\end{tabular} & \begin{tabular}[c]{@{}c@{}}North error \\ m\end{tabular} & \begin{tabular}[c]{@{}c@{}}Up error \\ m\end{tabular} & \begin{tabular}[c]{@{}c@{}}3D position \\ error m\end{tabular} & \begin{tabular}[c]{@{}c@{}}Total \\ computation\\ time ms\end{tabular} & \begin{tabular}[c]{@{}c@{}}Computation \\ time per \\ iteration ms\end{tabular} & \begin{tabular}[c]{@{}c@{}}Number of \\ iterations\end{tabular} \\ \hline
   \begin{tabular}[c]{@{}c@{}}GNSS \\ position\end{tabular}   & 1.003                                                   & 0.990                                                    & 0.997                                                 & 1.726                                                          & -                                                                      & -                                                                               & -                                                               \\
   6-DOF IMU                                                  & 0.293                                                   & 0.260                                                    & 0.310                                                 & 0.499                                                          & 629.9                                                                  & 105.0                                                                           & 6                                                               \\
   \begin{tabular}[c]{@{}c@{}}Proposed \\ method\end{tabular} & 0.327                                                   & 0.299                                                    & 0.265                                                 & 0.516                                                          & 237.0                                                                  & 33.9                                                                            & 7                                                               \\ \hline
   \end{tabular}%
   }
\end{table}

\begin{figure}[t]
   \centering
   \includegraphics[width=78mm]{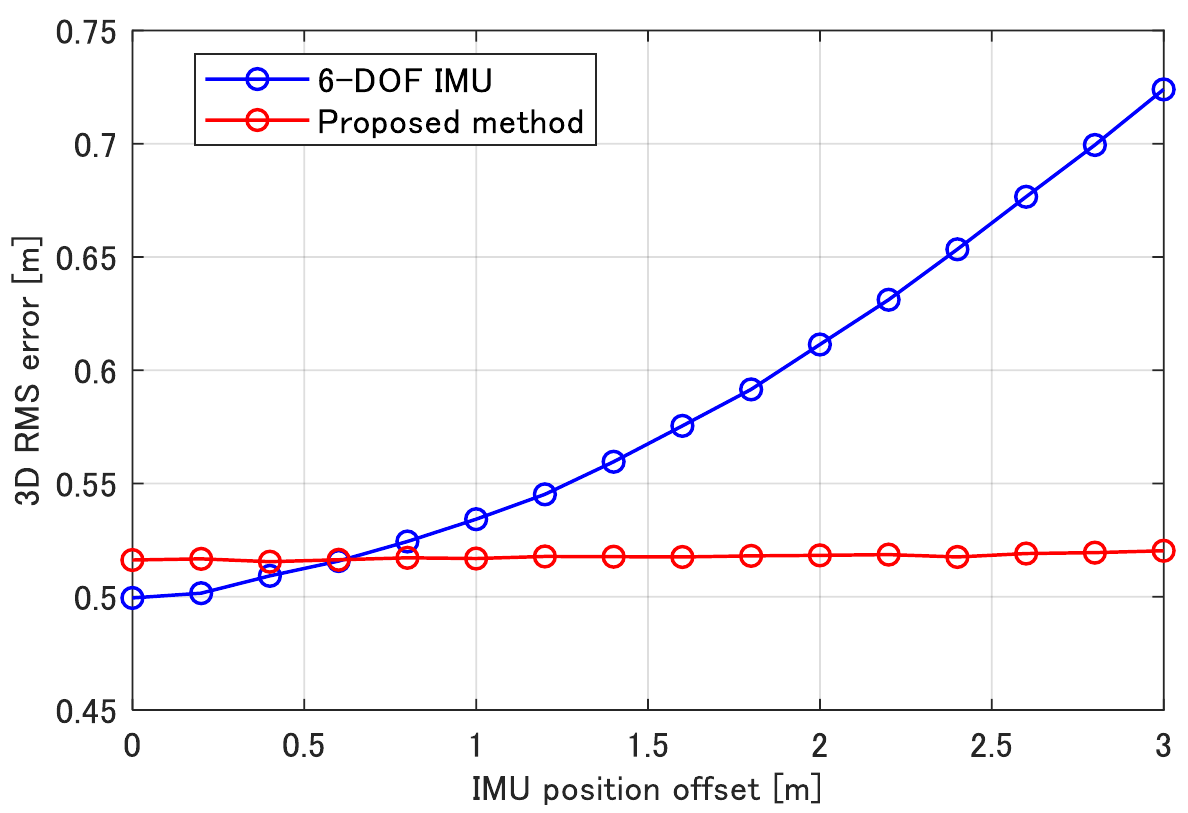}
   \caption{Comparison of position estimation error with increasing IMU mounting position error. The proposed method (red line) is less sensitive to the IMU mounting position error, while the conventional 6-DOF pose estimation (blue line) exhibits increasing error.}
   \label{fig5}
\end{figure}

\begin{figure}[t]
   \centering
   \includegraphics[width=78mm]{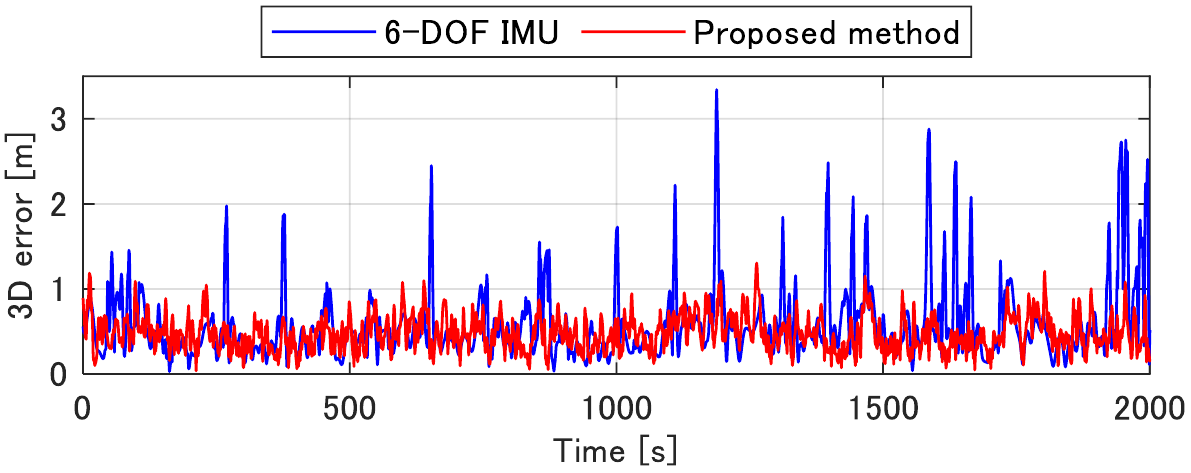}
   \caption{Comparison of 3D position error between the proposed method and the conventional method when the IMU mounting position error is 3 m. In the case of the conventional method, the position estimation error increases at points where the vehicle travels around curves.}
   \label{fig6}
\end{figure}

\begin{figure}[t]
   \centering
   \includegraphics[width=80mm]{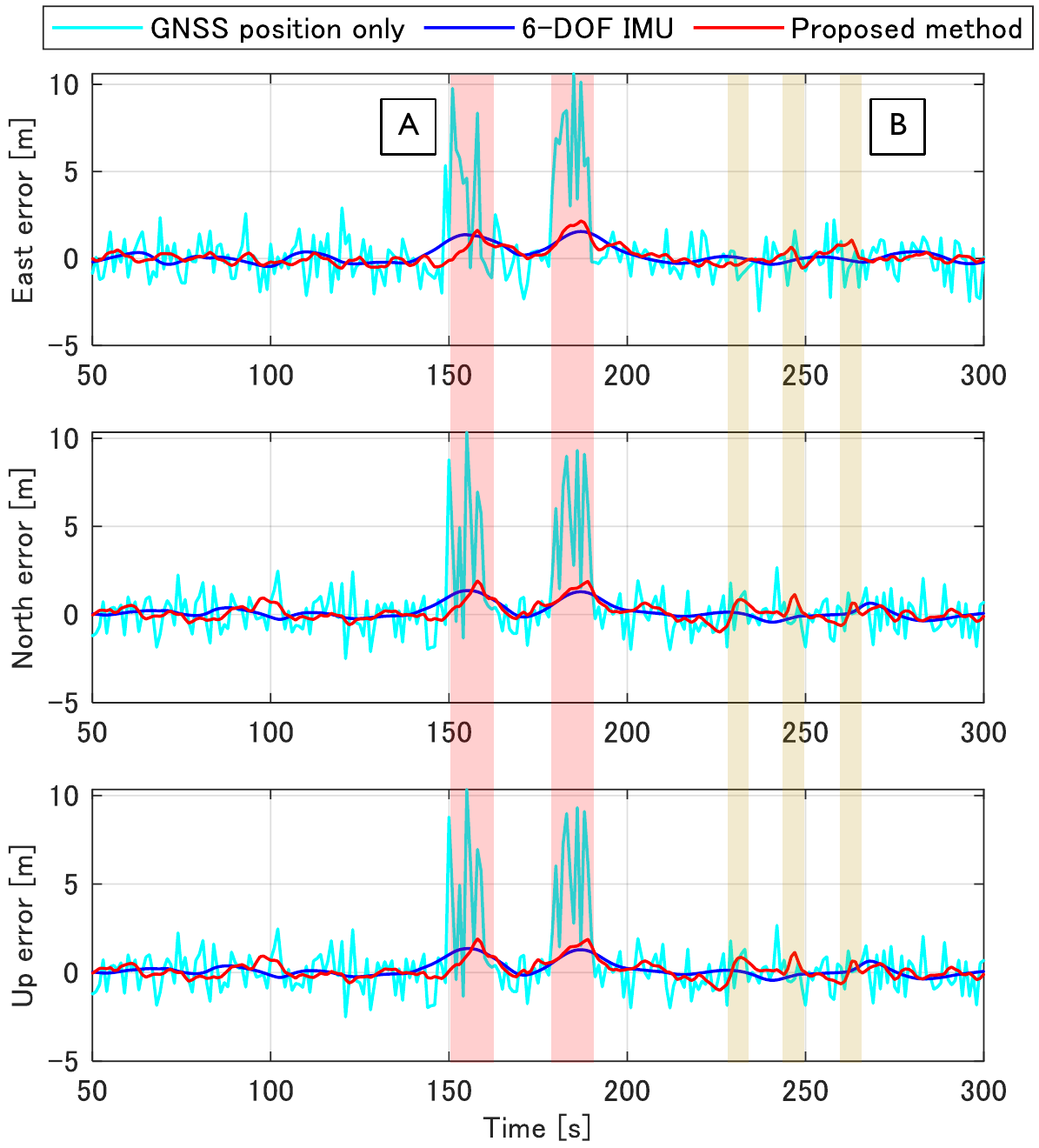}
   \caption{Position estimation errors of the proposed (red line) and conventional (blue line) methods when multipath is included in GNSS observations (A) and in the absence of GNSS observations(B).}
   \label{fig7}
\end{figure}

\subsection{Comparison using the IMU Mounting Position Error}
Fig. \ref{fig4} shows the cumulative distribution functions of the 3D position error when there is no IMU mounting position error for the proposed (red line) and conventional 6-DOF pose estimation (blue line). Table II lists the root mean square (RMS) error of the estimated positions, the total processing time for the optimization computation, the average processing time per iteration, and the number of iterations until convergence.

In the absence of the IMU mounting position error (when the IMU frame was coincident with the vehicle frame), the proposed method improved the estimation accuracy. However, the position estimation accuracy was slightly lower than that of the conventional method. This is because the conventional method can ideally integrate the GNSS and IMU when the sensor model parameters are known. In addition, the proposed method did not maximize the use of the three-axis observations of the IMU and limited the observations. This degraded the performance compared to the case where full acceleration and angular velocity observations were used. The proposed method is superior to the conventional method in terms of computational time because it estimates fewer states, which reduces the time required for an optimization iteration to about one third of the time required by the conventional method.

Fig. \ref{fig5} shows the RMS error in 3D position estimation when the simulation data is generated by adding a constant IMU mounting position error from the origin of the vehicle frame to the lateral direction of the vehicle. As shown in Fig. \ref{fig5}, the position estimation accuracy of the proposed method is less affected by the IMU mounting position. However, the conventional method, which is a 6-DOF pose estimation method, demonstrated that the accuracy of position estimation decreased with increasing IMU mounting position error. When the IMU mounting position error in the direction lateral to the vehicle exceeded 0.6 m, the proposed method exhibited better position estimation accuracy and surpassed the conventional method.

Fig. \ref{fig6} shows the time series of the 3D position errors for an IMU mounting position error of 3 m. The conventional 6-DOF attitude estimation method exhibits a partial increase in the position estimation error, which coincides with a vehicle traveling around a turn. When the IMU was mounted far from the center of rotation and generated a large centripetal force, the generated centripetal force directly affected the 3D attitude estimation, which in turn affected the position estimation accuracy. However, because the proposed method did not estimate the 3D attitude, the magnitude of the acceleration generated by the centripetal force affected the accuracy, although the effect was smaller than that of the conventional method, which estimates the 3D attitude.

\subsection{Comparison by Multipath and GNSS Shielding}
We evaluated the degradation of the position estimation accuracy of each method by simulating the cases where the GNSS observations exhibited large position and velocity errors owing to multipath and the satellite had a short time loss owing to shielding. Fig. \ref{fig7} shows the position estimation errors for the case without IMU, 6-DOF pose estimation, and the proposed method. The section "A" is a randomly observed position and velocity with errors added by simulating multipath errors. The magnitude of the multipath comprised random errors of up to 10 m and up to 1 m/s for position and velocity observations, respectively. In the absence of the IMU, the position estimation accuracy in section "A" was significantly degraded because of multipath. The combined use of the IMU and the proposed method significantly reduced the estimation error for both the proposed and conventional methods. 

In section "B,” the complete loss of GNSS observations for 5 seconds was repeated thrice. Even when GNSS signals were unavailable, the position was calculated using IMU. Both the proposed and conventional methods suppressed the increase in the position estimation error. The proposed method improved the accuracy of position estimation even when the GNSS observation data contained errors and deficiencies by combining GNSS and IMU without estimating the attitude.

\begin{figure*}[t]
   \centering
   \begin{minipage}{0.49\linewidth}
    \begin{center}
     \includegraphics[width=65mm]{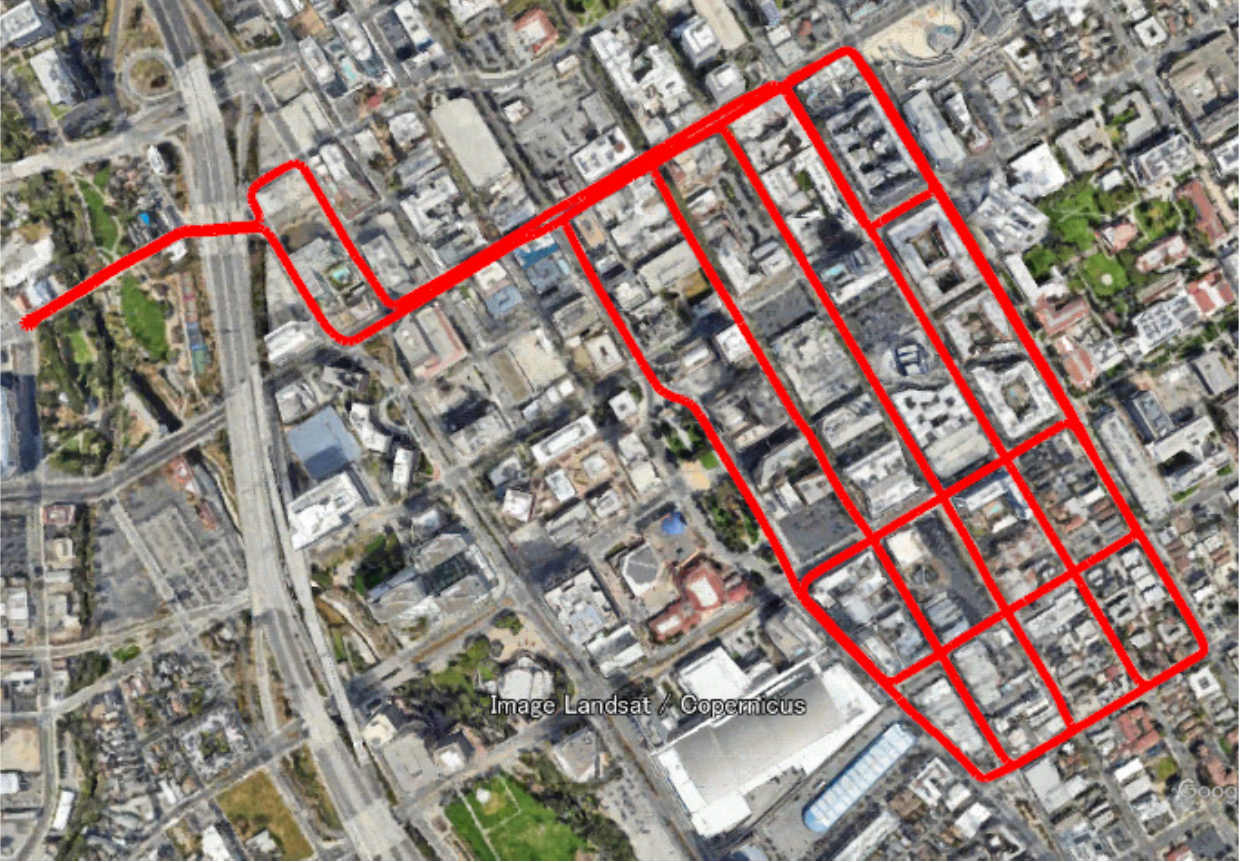}
    \end{center}
    \caption{Driving trajectory of a vehicle equipped with a smartphone. The experimental environment is a downtown area, where GNSS signal shielding and multipath occur frequently.}
    \label{fig8}
   \end{minipage}
   \hspace{0.005\linewidth}
   \begin{minipage}{0.49\linewidth}
    \begin{center}
     \includegraphics[width=65mm]{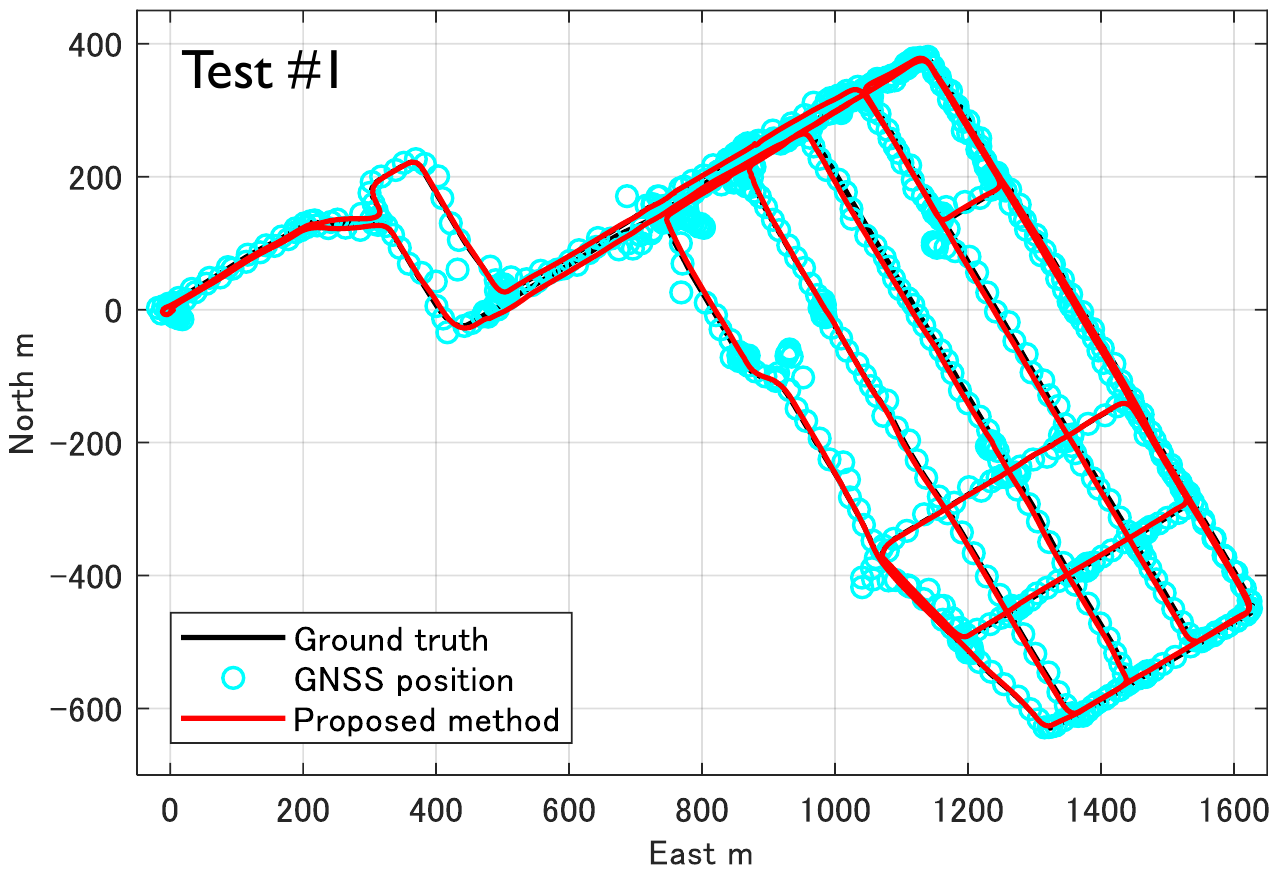}
    \end{center}
    \caption{Position estimation results using the proposed method, where the combined use of IMUs provides smooth position estimation regardless of GNSS errors and observation deficiencies.}
    \label{fig9}
   \end{minipage}
\end{figure*}
\begin{figure*}[t]
   \centering
   \includegraphics[width=155mm]{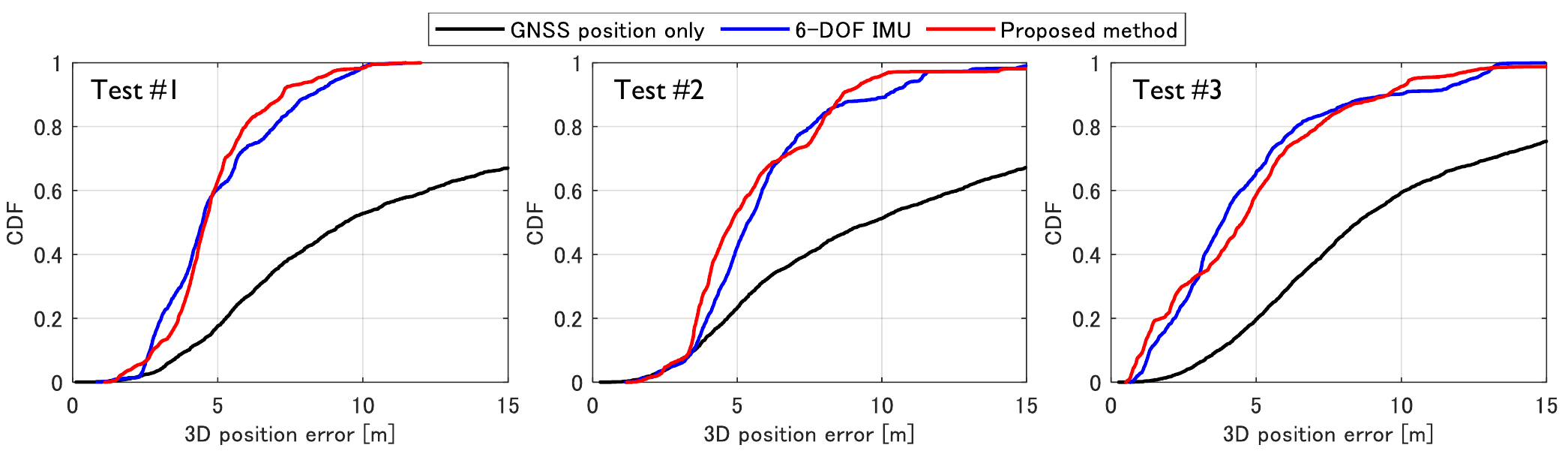} 
   \caption{Comparison of the cumulative distribution function of the proposed (red line) and conventional (blue line) methods for three vehicle driving experiments. The proposed method performs slightly better than the conventional method.}
   \label{fig10}
\end{figure*}

\begin{table*}[]
   \caption{RMS position error of the proposed method and conventional 6-DOF pose estimation for the actual smartphone dataset.}
   \label{tab3}
   \resizebox{\textwidth}{!}{%
   \begin{tabular}{@{}c|cccc|cccc|cccc@{}}
   \toprule
    & \multicolumn{4}{c|}{Test \#1} & \multicolumn{4}{c|}{Test \#2} & \multicolumn{4}{c}{Test \#3} \\ \midrule
    & East error m & North error m & Up error m & 3D error m & East error m & North error m & Up error m & 3D error m & East error m & North error m & Up error m & 3D error m \\
   GNSS position only & 8.671 & 11.228 & 23.808 & 27.714 & 9.698 & 10.495 & 22.782 & 26.893 & 11.920 & 9.179 & 15.506 & 21.605 \\
   6-DOF IMU & 3.674 & 2.448 & 3.133 & 5.413 & 4.752 & 3.050 & 3.239 & 6.510 & 3.976 & 2.860 & 2.804 & 5.643 \\
   Proposed method & 3.815 & 2.754 & 2.075 & 5.143 & 5.008 & 3.078 & 2.253 & 6.295 & 3.900 & 3.060 & 3.063 & 5.827 \\ \bottomrule
   \end{tabular}%
   }
\end{table*}

\subsection{Actual Smartphone Dataset}
The GNSS and IMU data collected in a real-world environment were used to evaluate the proposed method. The Google Smartphone Decimeter Challenge dataset \cite{gsd} was used for the evaluation. For this dataset, a smartphone was mounted on the vehicle dashboard and data were collected from its built-in GNSS and IMU. The mounting position and orientation of the smartphone were not provided. For the evaluation in this study, the smartphone was assumed to be installed at the center of rotation of the vehicle.

The vehicle trajectories were the same as those used to generate the simulation data, and multiple trajectories measured on different dates and times were evaluated. Fig. \ref{fig8} shows the driving trajectory of a vehicle equipped with a smartphone. The driving environment was an urban environment lined with buildings with large multipath errors, where GNSS was completely blocked for several seconds by tall structures and buildings. 

The IMU and GNSS of smartphones have the problem of observation time offset. In this study, fixed time offsets between the GNSS and IMU were manually estimated and applied \cite{imusync}. Parameters such as the IMU noise were manually tuned for both the proposed and conventional methods.

The GNSS 3D position observations were based on the position information contained in the dataset estimated by the weighted least-squares method using the pseudorange from the raw GNSS data of the smartphone. The 3D velocity data were estimated by the least squares method from the Doppler observations of the smartphone.

The proposed and conventional methods were compared for three sets of driving data collected on different dates. Fig. \ref{fig9} shows an example of a smartphone trajectory estimated by the proposed method. Although the GNSS position contains a very large positioning error owing to multipath, the proposed method estimates a smooth trajectory by combining IMUs. Fig. \ref{fig10} compares the cumulative distribution functions of the proposed (red line) and conventional (blue line) methods for three vehicle driving experiments. Table III lists the RMS errors for the 3D position estimation. By combining the IMUs, both the proposed and conventional methods estimated a continuous smooth position without increasing the error, even in areas where the GNSS pseudorange-based position had large errors in the observations. The proposed and conventional methods exhibited almost the same level of position-estimation performance. However, the proposed method performed better than the conventional method for some runs. This may be attributed to the influence of mounting errors and degradation of the IMU data acquired in the real environment by noise and vibration. The proposed method greatly improves the accuracy of vertical position estimation. This is because for a ground moving vehicle, the direction of the acceleration vector occurs mainly in the horizontal direction, and the vertical velocity is more easily corrected than the horizontal velocity under the 3D position constraint when the magnitude constraint of the acceleration vector is used.

\section{Conclusion}
In this study, we proposed a combined GNSS and IMU method that does not require attitude estimation. In the state estimation problem by graph optimization, we proposed a constraint between states using the magnitude of the 3D acceleration vector observed by the IMU and a constraint on the angle of the velocity vector between states using the angular change using the gyroscope observation values. We conducted an evaluation using simulation data and smartphone observations in a real environment. In the evaluation using simulation data, the proposed method improved the position estimation accuracy even when there were large errors or deficiencies in the GNSS observations. The proposed method was not significantly affected by the IMU mounting position and the position estimation accuracy did not deteriorate when there were errors in the IMU mounting position. In the real-world position estimation using smartphones, the proposed method improved the positioning accuracy by combining IMUs. In conclusion, the proposed method was highly effective in the combined GNSS and IMU scenario where the IMU was frequently installed and removed.

In this study, we evaluated the performance of the proposed method by integrating the loosely coupled GNSS position and velocity with the IMU. In future studies, we will improve the position estimation performance by evaluating the combination of the IMU with a tight-coupling GNSS pseudorange and Doppler, and by including advanced velocity information such as the time-differenced carrier phase.

\ifCLASSOPTIONcaptionsoff
  \newpage
\fi



\bibliographystyle{IEEEtran}
\bibliography{IEEEabrv,RAL2023}

\end{document}